\documentclass[letterpaper, 10 pt, journal, twoside]{IEEEtran} 

\usepackage{graphics} 
\usepackage{epsfig} 
\usepackage{mathptmx} 
\usepackage{times} 
\usepackage{amsmath} 
\usepackage{amssymb}  

\usepackage[dvipsnames]{xcolor}

\usepackage{times}
\usepackage{epsfig}
\usepackage{graphicx}
\usepackage{amsmath}
\usepackage{amssymb}
\usepackage{array}
\newcolumntype{V}{>{\centering\arraybackslash} m{.5\linewidth} }
\newcolumntype{C}[1]{>{\centering\let\newline\\\arraybackslash\hspace{0pt}}m{#1}}
\usepackage{tabularx}
\newcolumntype{Y}{>{\centering\arraybackslash}X}
\usepackage{rotating}
\usepackage{multirow, makecell}
\usepackage{booktabs}
\usepackage{hyperref}
\hypersetup{
    colorlinks=true,
    linkcolor=blue,
    filecolor=magenta,      
    urlcolor=cyan,
}
\usepackage{subcaption}

\usepackage{dblfloatfix}

\title{
Fusing Body Posture with Facial Expressions for Joint Recognition of Affect in Child-Robot Interaction
}

\author{Panagiotis P. Filntisis$^{\:1,3}$, Niki Efthymiou$^{\:1,3}$, Petros Koutras$^{\:1,3}$, Gerasimos Potamianos$^{\:2,3}$ and Petros Maragos$^{\:1,3}$ \\ $^1$School of E.C.E., NTUA, Greece \quad  $^2$E.C.E. Department, UTH, Greece \quad $^3$Athena RC, Maroussi, Greece\\
\tt\small{\{filby, nefthymiou\}@central.ntua.gr,\{pkoutras,maragos\}@cs.ntua.gr,gpotam@ieee.org
}
}

\begin{document}

\markboth{IEEE Robotics and Automation Letters. Preprint Version. Accepted July, 2019}
{Filntisis \MakeLowercase{\textit{et al.}}: Fusing Body Posture with Facial Expressions for Joint Recognition of Affect in CRI}  

\author{Panagiotis P. Filntisis$^{1,3}$, Niki Efthymiou$^{1,3}$, Petros Koutras$^{1,3}$, Gerasimos Potamianos$^{2,3}$ and Petros Maragos$^{1,3}$

\thanks{This paper was recommended for publication by Editor Dongheui Lee upon evaluation of the Associate Editor and Reviewers' comments. 
This work was supported by the EU-funded Project BabyRobot (H2020 under Grant Agreement
687831).} 
\thanks{$^{1,3}$P.P. Filntisis, N. Efthymiou, P. Koutras and P. Maragos are with the School of Electrical and Computer Engineering, National Technical University of Athens, Zografou 15780, Greece and with the Athena Research and Innovation Center, Marousi 15125, Greece (email: 
        {\tt\footnotesize \{filby,nefthymiou\}@central.ntua.gr,\{pkoutras,maragos\}  @cs.ntua.gr}).}%
\thanks{$^{2,3}$G. Potamianos is with the Department of Electrical and Computer Engineering, University of Thessaly, Volos 38221, Greece and with the Athena Research and Innovation Center, Marousi 15125, Greece
        (email: {\tt\footnotesize gpotam@ieee.org}).}%


\thanks{Digital Object Identifier (DOI): \url{10.1109/LRA.2019.2930434}}
\thanks{\textcopyright 2019 IEEE.  Personal use of this material is permitted.  Permission from IEEE must be obtained for all other uses, in any current or future media, including reprinting/republishing this material for advertising or promotional purposes, creating new collective works, for resale or redistribution to servers or lists, or reuse of any copyrighted component of this work in other works.}
}



\maketitle


\begin{abstract}
In this paper we address the problem of multi-cue affect recognition in challenging scenarios such as child-robot interaction. Towards this goal we propose a method for automatic recognition of affect that leverages body expressions alongside facial ones, as opposed to traditional methods that typically focus only on the latter. Our deep-learning based method uses hierarchical multi-label annotations and multi-stage losses, can be trained both jointly and separately, and offers us computational models for both individual modalities, as well as for the whole body emotion. We evaluate our method on a challenging child-robot interaction database of emotional expressions collected by us, as well as on the GEMEP public database of acted emotions by adults, and show that the proposed method achieves significantly better results than facial-only expression baselines.

\end{abstract}



\section{Introduction}
\IEEEPARstart{S}{ocial} robotics is a fairly new area in robotics that has been enjoying a swift rise in its applications, some of which include robot assisted therapy in adults and children~\cite{belpaeme2013child}, activities of daily living~\cite{broadbent2009acceptance}, and education~\cite{belpaeme2018social}. A critical capability of social robots is empathy: the capacity to correctly interpret the social cues of humans that are manifestations of their affective state. Empathic agents are able to change their behavior and actions according to the perceived affective states and as a result establish rapport, trust, and healthy long-term interactions \cite{bickmore2005establishing}. Especially in the field of education, empathic robot behaviors that are congruent with the child's feelings increase trust and have a positive impact to the child-robot relationship, whereas incongruent behavior has a significantly negative effect~\cite{leite2014empathic}.

An important factor in many social robot applications, and especially in child-robot interaction (CRI) \cite{tsiami2018multi3}, is the fact that the flow of interaction is unpredictable and constantly fluctuating~\cite{ros2011child}. Although interaction with adults can usually be restricted and controlled, the spontaneous nature of children fails to meet this criterion and becomes a true challenge. A direct implication is the fact that robots can no longer rely only on facial expressions to recognize emotion, which is the main visual cue employed in automatic affect recognition~\cite{de2009bodies}, but also have to take into account body expressions that can stay visible and detectable even when the face is unobservable.

\begin{figure}[t]
   \includegraphics[width=1\linewidth]{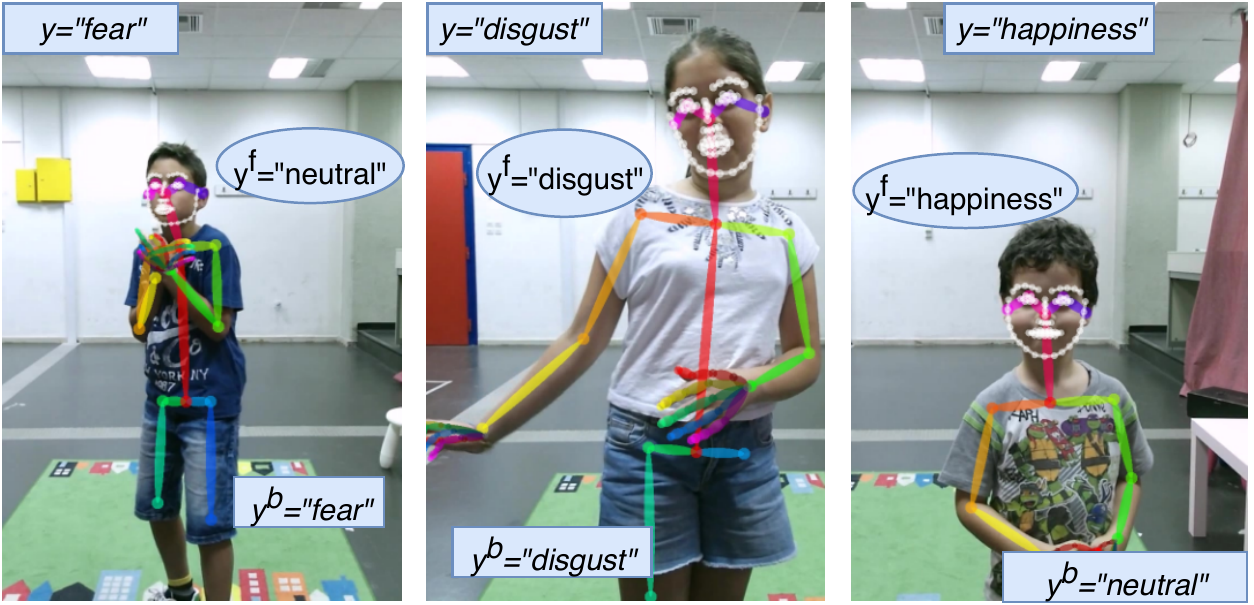}
    \caption{Hierarchical multi-labels for affect recognition via body and face cues, where $y$ denotes the whole body emotion label, $y^f$ the facial expression label, and $y^b$ the body expression one.}
    \label{fig:f}
    \vspace{-0.5cm}
\end{figure}

Research in bodily expression of emotion suggests that emotion is equally conveyed through bodily expressions and actions in most cases \cite{de2009bodies, wallbott1998bodily}, while both the static body posture as well as the dynamics \cite{atkinson2004emotion,calvo2015oxford} contribute in its perception. Furthermore, there are emotions such as pride \cite{tracy2004show} that are more discernible through body rather than face observation. An also consistent finding in multiple studies is the fact that considering both body and face concurrently increases emotion recognition rates \cite{van2007body}. Aviezer et al. also point out in \cite{aviezer2012body} that the body can be a deciding factor in determining intense positive or negative emotions. However, to date, most research has focused on automatic visual recognition of emotion from facial expressions \cite{jung2015joint,Kuo_2018_CVPR_Workshops}, with only few including emotional body expressions into the recognition loop \cite{de2009bodies}.

\textcolor{black}{Motivated by the above, in this paper, we propose an end-to-end system of automatic emotion recognition for CRI that hierarchically fuses body and facial features. We show that by using only a skeleton structure which describes the human pose obtained by leveraging the latest advancements in human pose recognition \cite{cao2017realtime}, we can satisfactorily assess the human emotion and increase the performance of automatic affect recognition in CRI scenarios.} In summary, our contributions are as follows:

\begin{itemize}
    \item We propose a method based on Deep Neural Networks (DNNs) that fuses body posture skeleton information with facial expressions for automatic recognition of emotion. The networks can be trained both separately and jointly, and result in significant performance boosts when compared to facial-only expression baselines.
    \item We use hierarchical multi-label annotations (Figure~\ref{fig:f}), that describe not only the emotion of the person as a whole, but also the separate body and facial expressions. These annotations allow us to train, either jointly or separately, our hierarchical multi-label method, providing us with computational models for the different modalities of expressions as well as their fusion.
    \item We develop and analyze a database containing acted and spontaneous affective expressions of children participating in a CRI scenario, and we discuss the challenges of building an automatic emotion recognition system for children. The database contains emotional expressions both in face and posture, allowing us to observe and automatically recognize patterns of bodily emotional expressions across children in various ages. 
\end{itemize}

The remainder of the paper is organized as follows: Section \ref{sec:2} presents previous works in emotion recognition based on facial and body expressions, as well as related research on applications of emotion recognition in CRI scenarios. In Section \ref{sec:3} we present our method for automatic recognition of affect by fusing body posture and facial expressions with hierarchical multi-label training. Section \ref{sec:4} describes the BabyRobot Emotion Database that has been collected for evaluating our approach. Then, Section \ref{sec:5} includes our experimental results, and Section \ref{sec:6} our conclusive remarks and future directions.

\section{Related Work}
\label{sec:2}
The overwhelming majority of previous works in emotion recognition from visual cues have focused on using only facial information \cite{de2009bodies}. Recent surveys however \cite{noroozi2018survey,Karg13,kleinsmith13} highlight the need for taking into account bodily expression as additional input to automatic emotion recognition systems, as well as the lack of large-scale databases for this task.

Gunes and Piccardi \cite{gunes2009automatic} focused on combining handcrafted facial and body features for recognizing 12 different affective states in a subset of the FABO database~\cite{gunes2006bimodal} that contains upper body affective recordings of 23 subjects. Barros et al.~\cite{barros2015multimodal} used Sobel filters combined with convolutional layers on the same database, while Sun et al.~\cite{sun2018affect} employed a hierarchical combination of bidirectional long short-term memory (LSTM) and convolutional layers for body-face fusion using support vector machines. \textcolor{black}{Piana et al. \cite{Piana:2016:ABG:2896319.2818740} built an automatic emotion recognition system that exploits 3D human pose and movements and explored different higher level features in the context of serious games for autistic children.}

B\"{a}nziger et al.~\cite{banziger2012introducing} introduced the GEMEP (GEneva Multimodal Emotion Portrayal) corpus, the core set of which includes 10 actors performing 12 emotional expressions. In~\cite{dael2012body}, Dael et al. proposed a body action and posture coding system similar to the facial action coding system \cite{friesen1978facial}, which is used for coding human facial expressions, and subsequently utilized it in \cite{dael2012emotion} for classifying and analyzing body emotional expressions found in the GEMEP corpus.

In \cite{castellano2008emotion}, Castellano et al. recorded a database of 10 participants performing 8 emotions, using the same framework as the GEMEP dataset. Afterwards, they fused audio, facial, and body movement features using different Bayesian classifiers for automatically recognizing the depicted emotions. In \cite{psaltis2016multimodal}, a two-branch face-body late fusion scheme is presented by combining handcrafted features from 3D body joints and action units detection using facial landmarks.

Regarding the application of affect recognition in CRI, the necessity of empathy as a primary capability of social robots for the establishment of positive long-term human-robot interaction has been the research focus of several studies \cite{bickmore2005establishing,leite2014empathic}. In ~\cite{castellano2013multimodal}, Castellano et al. presented a system that learned to perceive affective expressions of children playing chess with an iCat robot and modify the behavior of the robot resulting in a more engaging and friendly interaction. An adaptive robot behavior based on the perceived emotional responses was also developed for a NAO robot in \cite{tielman2014adaptive}. In \cite{marinoiu20183d}, 3D human pose was used for estimating the affective state of the child in the continuous arousal and valence dimensions, during the interaction of autistic children with a robot.

\textcolor{black}{Compared to the existing literature, our work introduces hierarchical multi-labels, by taking into account the medium through which a person expresses its emotion (face and/or body). These labels are used in a novel neural network architecture that utilizes multi-stage losses, offering tighter supervision during training, as well as different sub-networks, each specialized in a different modality. Our method is end-to-end, uses only RGB information, and is built with the most recent ML architectures. The efficiency of the proposed framework is validated by performing extensive experimental results on two different databases, one of which includes emotions acted by children and was collected by us during the EU project BabyRobot\footnote{More info: \href{http://babyrobot.eu/}{http://babyrobot.eu/}}.}

\begin{figure*}[t]
    \centering
    \includegraphics[width=1\linewidth]{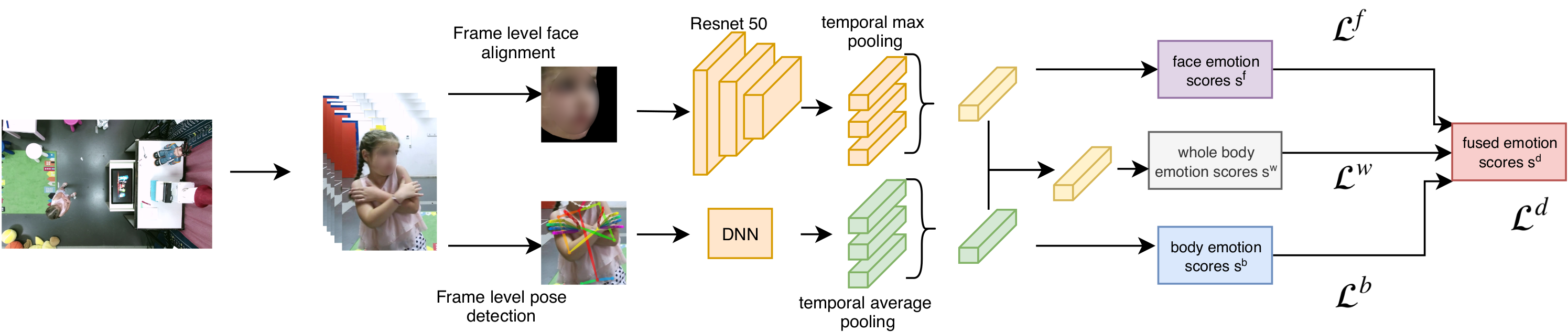}
    \caption{Hierarchical multi-label training for recognition of affect from multiple visual cues in CRI.}
    \label{fig:arch}
            \vspace{-0.5cm}
\end{figure*}

\section{Whole Body Emotion Recognition}
\label{sec:3}
In this section we first present an analysis of bodily expression of emotion. Then, we detail our method for automatic recognition of affect.

\subsection{Bodily expression of emotion}
While the face is the primary medium through which humans express their emotions (i.e., an affect display \cite{ekman1967head}), in real life scenarios it is more often that we find ourselves decoding the emotions of our interlocutor or people in our surroundings by observing their body language, especially in cases where either the face of the subject in question is occluded, hidden, or far in the distance. In general, the body language can act both as a supportive modality, in which case it enforces the confidence on an already recognized emotion from the face or provides crucial missing information (e.g., in cases where the face cannot reliably identify the emotion due to its intensity \cite{aviezer2012body}), as well as a primary modality, in which case it is the only source of information from which we can deduce the emotion.


A problem that arises when dealing with spontaneous (i.e., not acted) or in-the-wild data is the fact that different individuals express themselves through different modalities, depending on which cue they prefer using (body, face, voice) \cite{ExpressingEmotionThroughPostureandGesture}. This fact is cumbersome for supervised learning algorithms, e.g., in samples where an emotion label corresponds to the facial expression only and not the body, which means that the subject in question preferred to use only the face while the body remained neutral. In such data, one way to alleviate this issue is to include hierarchical labels, which first denote the ground truth labels of the different modalities.
Examples of hierarchical multi-labels are shown in Figure~\ref{fig:f}, where $y$ denotes the emotion the human is expressing (which we call the ``whole" body label), $y^f$ the emotion that is conveyed through the face (i.e., $y^f = y$ if the subject uses the face, else $y^f = ``\text{neutral}"$), and $y^b$ the emotion that is conveyed through the body (i.e., $y^b = y$ if the subject uses the body, else $y^b = ``\text{neutral}"$).


\subsection{Method} 
\label{sec:method}
Based on the aforementioned analysis, Figure \ref{fig:arch} presents our DNN architecture for automatic multi-cue affect recognition using hierarchical multi-label training (HMT). We assume that we have both the whole body label $y$, as well as the hierarchical labels  $y^f$ for the face and $y^b$ for the body. \textcolor{black}{The network initially consists of two different branches, with one branch focusing on facial expressions, and one branch focusing on body posture. The two branches are then combined at a later stage to form the whole body expression recognition branch that takes into account both sources of information. This design allows setting up different losses on different stages of the network based on the hierarchical labels, offering stricter supervision during training. The output of the network is the recognized emotional state of the person detected in the input video.} 


\paragraph{Facial Expression Recognition Branch}
The facial expression recognition branch of the network is responsible for recognizing emotions by decoding facial expressions. If we consider a frame of a video sequence ${I_i}\lvert_{i=1,..,N}$, at each frame we first apply a head detection and alignment algorithm in order to obtain the cropped face image (see Section \ref{sec:crop_face}). This is subsequently fed into a Residual Network \cite{he2016deep} CNN architecture to get a 2048-long feature vector description of each frame ${H_i^f}\lvert_{i=1,..,N}$. Then, we apply temporal max pooling over the video frames to obtain the representation of the facial frame sequence: 
\begin{equation}
    H^f = \max_i{H^f_i}\lvert_{i=1,...,N}
\end{equation}
\textcolor{black}{By assuming that the feature map obtains its maximum values in frames where the facial expression is at peak intensity, max pooling selects only the information regarding the facial expressions at their peak over the frame sequence.} Then, we apply a fully connected (FC) layer on $H^f$ to obtain the facial emotion scores, $s^f$. 

We can calculate the loss obtained through this branch as the cross entropy $\mathcal{L}^f (y^f,\tilde{s}^f)$ between the face labels $y^f$ and the probabilities of the face scores $\tilde{s}^f$ obtained via a softmax function:
\begin{equation}
\mathcal{L}^f (y^f,\tilde{s}^f) = -\sum^{C}_{c=1} y^f_c \log \tilde{s}^f_c
\label{eq:ce}
\end{equation}
with $C$ denoting the number of emotion classes.
\paragraph{Body Expression Recognition Branch}
In the second branch, for each frame of the input video ${I_i}\lvert_{i=1,..,N}$, we apply a 2D pose detection method in order to get the skeleton $J_i \in \mathbb{R}^{K\times2}$, where $K$ is the number of joints in the detected skeleton (see Section \ref{sec:crop_face}). The 2D pose is then flattened and input as a vector into a DNN in order to get a representation $H^b_i\lvert_{i=1,..,N}$. We then apply global temporal average pooling (GTAP) over the entire input sequence:
\begin{equation}
    H^b = \frac{1}{N} \sum_{i=1}^{N} H_i^b
\end{equation}
\textcolor{black}{In contrast to the face branch, we use temporal average pooling for the body branch in order to capture the general pattern of the features during the temporal sequence and not completely discard temporal information.}
The scores for the body emotion $s^b$ are obtained by passing the pose representation of the video $H^b$ over an FC layer. The loss in this branch is the cross entropy loss (Eq. \ref{eq:ce}) between the body labels $y^b$ and the probabilities $\tilde{s}^b$, $\mathcal{L}^b (y^b,\tilde{s}^b)$.

\paragraph{Whole Body Expression Recognition Branch}
In order to obtain whole body emotion recognition scores $s^w$, we concatenate $H^f$ and $H^b$ and feed them through another FC. We then use the whole body emotion labels $y$ to obtain the whole body cross entropy loss between the whole body labels $y$ and the probabilities $\tilde{s}^w$, $\mathcal{L}^w (y,\tilde{s}^w)$.

\paragraph{Fusion}
Finally, we employ a fusion scheme as follows: we concatenate the scores $s^f$, $s^b$, and $s^w$ and use a final FC in order to obtain the fused scores $s^d$. This way we get a final loss $\mathcal{L}^d (y,\tilde{s}^d)$ which is the cross entropy between the whole body labels $y$ and $\tilde{s}^d$.

During training, the loss that is backpropagated through the network is:
\begin{equation}
    \mathcal{L} = \mathcal{L}^f (y^f,\tilde{s}^f) + \mathcal{L}^b (y^b,\tilde{s}^b) + \mathcal{L}^w (y,\tilde{s}^w) + \mathcal{L}^d (y,\tilde{s}^d)
\end{equation}
The network final prediction of the human affect in the video is obtained by the fusion score vector $s^d$.

\section{The BabyRobot Emotion Database}
\label{sec:4}
\begin{figure}[t]
\centering
  \includegraphics[width=1\linewidth]{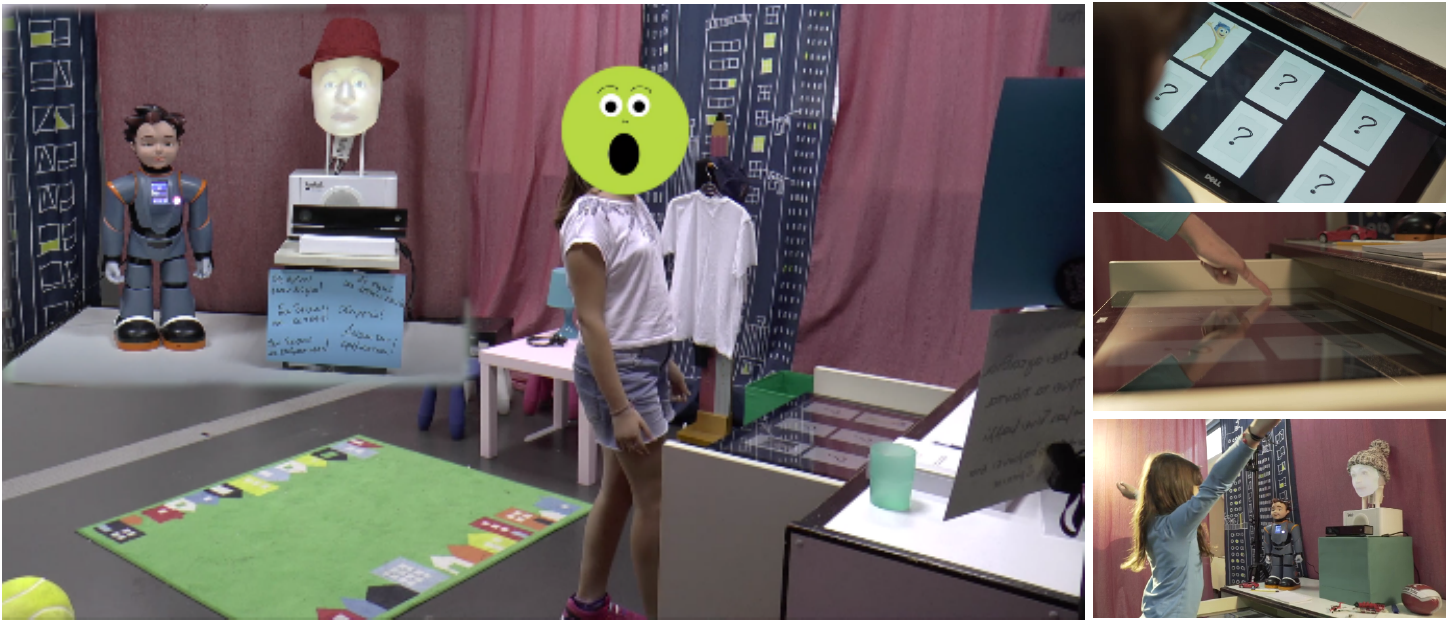}
        \caption{The experimental setup of the BRED Database and snapshots showing children playing the ``Express the feeling" game.}
        \label{fig:cri}
        \vspace{-0.6cm}
\end{figure}

In order to evaluate our method, we have collected a database which includes multimodal recordings of children interacting with two different robots (Zeno \cite{zeno}, Furhat \cite{furhat}), in a laboratory setting that has been decorated in order to resemble a child's room (Figure \ref{fig:cri}). 

We call this dataset the BabyRobot Emotion Database (BRED). BRED includes two different kinds of recordings: \textcolor{black}{\textit{Pre-Game Recordings}} during which children were asked \textcolor{black}{by a human} to express one of six emotions, and \textcolor{black}{\textit{Game Recordings}} during which children were playing a game called ``Express the feeling" with the Zeno and Furhat robots. The game was touchscreen-based, and throughout its duration children selected face-down cards, each of which represented a different emotion. After seeing the cards, the children were asked to express the emotion, and then one of the robots followed up with a facial gesture that expressed the emotion as well. A total of 30 children of ages between 6 to 12 took part in \textcolor{black}{both} recordings. It is important to note that we did \textit{not} give any guidelines or any information to the children on how to express their emotions. \textcolor{black}{The experimental procedure was approved by an Independent Ethics Committee from the Athena Research and Innovation Center in Athens, Greece.}

The emotions included in the database are: \textit{Anger}, \textit{Happiness}, \textit{Fear}, \textit{Sadness}, \textit{Disgust}, and \textit{Surprise}, the $6$ basic emotions included in Ekman and Freisen's initial studies \cite{ekman1971constants}. \textcolor{black}{This categorical representation of emotion is the most commonly used in research studies of automatic emotion recognition \cite{gunes2010automatic}, and is typically adopted across different databases of emotional depictions \cite{li2018deep}. When compared to dimensional approaches (e.g., valence/arousal space), the categorical emotional approach is less flexible in expressing more complex emotions, however it is easier to annotate \cite{gunes2013categorical}.}

\paragraph{Hierarchical Database Annotations} In total, the initial recordings included $180$ samples of emotional expressions from the ``Pre-Game" session and $180$ samples from the ``Game" session ($30$ children $\times$ $6$ emotions for both sessions). The annotation procedure included three different phases. In the first phase, $6$ different annotators filtered out recordings where the children did not perform any emotion (due to shyness, lack of attention, or other reasons), and identified the temporal segments during which the expression of emotions takes place (starting with the onset of the emotion and ending just before the offset). In the second phase, 2 annotators validated the annotations of the previous phase. \textcolor{black}{Finally in the third phase, three} different annotators annotated the videos hierarchically, by indicating for each video whether the child was using the face, body, or both, to express the emotion. \textcolor{black}{The final hierarchical labels were obtained using majority voting over the three annotations. Inter-annotator agreement was also measured using Fleiss' kappa coefficient \cite{fleiss1973equivalence}, with value $0.48$ for the face labels and $0.84$ for the body labels. The values show that for the body labels we have an almost perfect agreement between the annotators, while for the face labels there are some cases where the annotators disagreed due to really slight facial expressions.}

\begin{table}[t]
\centering

\begin{tabularx}{\linewidth}{|l|Y|Y|Y}
\hline
\textbf{Emotion}  & \textbf{\% using facial exp.} & \textbf{\% using body exp.} \\ \hline \hline
\textbf{Happiness}                       & 100\%                              & 20\%                               \\ 
\textbf{Sadness}                         & 86\%                              & 49\%                              \\ 
\textbf{Surprise}                        & 100\%                              & 43\%                               \\
\textbf{Fear}                            & 42\%                               & 98\%                              \\ 
\textbf{Disgust}                         & 98\%                              & 42\%                              \\ 
\textbf{Anger}                           & 85\%                              & 70\%                              \\  \hline
\end{tabularx}
\caption{Hierarchical multi-label annotations of the BabyRobot Emotion Dataset (BRED) depicting usage of body and facial expressions for each emotion.}
\label{tab:stats-ann}
\vspace{-0.6cm}

\end{table}

\begin{table*}[h]

\centering
\begin{tabular}{C{0.15\linewidth}|C{0.15\linewidth}|C{0.15\linewidth}|C{0.15\linewidth}|C{0.15\linewidth}|C{0.15\linewidth}}
\textbf{happiness} & \textbf{sadness} & \textbf{surprise} & \textbf{fear} & \textbf{disgust} & \textbf{anger}    \\
mainly facial, rare jumping and/or open raised hands, body erect, upright head & crying (hands in front on face), motionless, head looking down, contracted chest
& expanded chest, hand movement without specific patterns, either positive or negative & quick eye gaze, weak facial expressions, arms crossed in front of body, head sink & 
mainly facial (tongue out), movement away from/hands against robot &
clenched fists, arms crossed, squared shoulders 
\\
\includegraphics[width=1\linewidth]{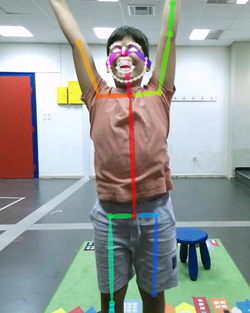}
  &
\includegraphics[width=1\linewidth]{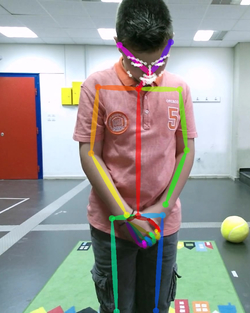}
&
\includegraphics[width=1\linewidth]{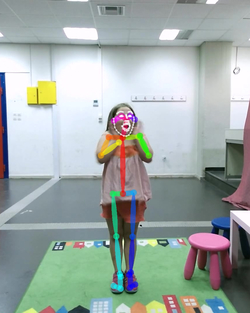}
&
\includegraphics[width=1\linewidth]{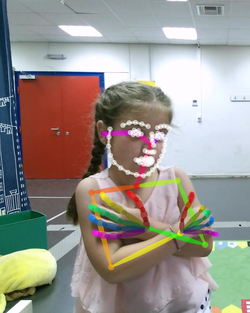}
  &
\includegraphics[width=1\linewidth]{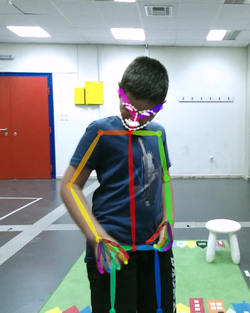}
&
\includegraphics[width=1\linewidth]{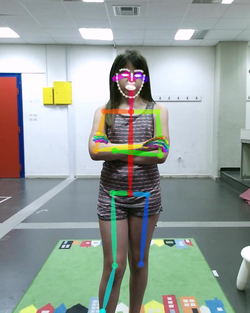}
\\

\end{tabular}
\caption{Patterns of bodily expression of emotion in the BRED corpus and example images.}
\label{tab:bodily_expression}
\end{table*}
\textcolor{black}{In total, the database features 215 valid emotion sequences, with an average length of 72 frames at 30FPS. The smaller number of valid sequences extracted from the 360 initial recordings shows that, when collecting data from children, attention should be paid in data validation and cleaning. Table \ref{tab:stats-ann} contains more insights regarding the database and its annotations. For each different emotion, we show the percentage of samples where the child used its face/body to depict emotion, against the number of total samples. We observe that almost all children used their body to express fear ($98\%$), while less than half used their face. Another emotion where a large percentage of children utilized the body is anger ($70\%$). To indicate happiness, surprise, and disgust, almost \textit{all} children used facial expressions ($100\%, 100\%,$ and $98\%$, respectively).}
Table \ref{tab:bodily_expression} also contains some of the annotators observations regarding the bodily expression of emotion in BRED, as well as examples from the database. All images include facial landmarks (although we do not use them in any way in our method) in order to protect privacy.

The newly collected BRED dataset is very challenging as it features many intra-class variations, multiple poses, and in many cases similar body expressions for different classes. \textcolor{black}{These include the similar pattern of hand crossing in anger and fear, and lowering of the head pose in fear and sadness. The BRED dataset is available at \href{https://zenodo.org/record/3233060}{https://zenodo.org/record/3233060}}.

\section{Experiments}
\label{sec:5}
In this section we present our experimental procedure and results. We first perform an exploratory analysis of the different branches and pathways of the HMT architecture of Figure \ref{fig:arch} on the GEMEP (GEneva Multimodal Emotion Portrayals) database \cite{banziger2012introducing}. As far as we are aware, this is the only publicly available video database that includes annotated whole body expressions of emotions. We believe that databases of upper body depictions, such as FABO \cite{gunes2006bimodal} where the subjects are sitting, restrict body posture expression and force the subjects to focus mostly on using their hands. Our main evaluation is then conducted on BRED where we experiment with variations of the HMT network.


\subsection{Network Setup and Initialization}
\label{sec:crop_face}

In order to avoid overfitting due to the small number of sequences in both GEMEP and BRED, and especially in the facial branch which includes a large number of parameters, we pretrain the branch on the AffectNet Database \cite{affectnet}. The AffectNet Database contains more than 1 million images of faces collected from the internet and annotated with one of the following labels: Neutral, Happiness, Anger, Sadness, Disgust, Contempt, Fear, Surprise, None, Uncertain, and Non-face. The manually annotated images amount to $440k$ with about $295k$ falling into one of the emotion categories (neutral plus 7 emotions). The database also includes a validation set of 500 images for each class, while the test set is not yet released.

\textcolor{black}{To prepare the facial branch for the subsequent feature extraction for our task, we start with a Resnet-50 CNN which has been trained using the ImageNet Database\footnote{These weights are provided by the PyTorch Framework. More information can be found in \href{https://pytorch.org/docs/stable/torchvision/models.html}{https://pytorch.org/docs/stable/torchvision/models.html}.}. Next, in order to learn features that are pertinent to our task, we train again the network, this time on AffectNet by replacing the final FC layer of the network with a new FC layer with 8 output classes (the 7 emotions of AffectNet plus neutral). The network was trained for 20 epochs using a batch size of 128 and the Adam optimizer \cite{adam}, achieving the best accuracy on the AffectNet validation set at the 13th epoch ($52.2\%$).} \textcolor{black}{As opposed to the facial branch, the body branch was not pretrained and its weights were initialized as in \cite{lecun2012efficient}.}

For detecting, cropping, and aligning the face for each frame, we use the OpenFace 2 toolkit \cite{openface}. We then use our pretrained facial branch to extract a $2048$-dimensional feature vector which is used during training. This means that during training the parameters of the feature extraction layers of the facial branch remain fixed.
Similarly, we extract the 2D pose of the subjects in \textcolor{black}{each database (GEMEP and BRED)} using OpenPose \cite{cao2017realtime} along with the 2D hand keypoints \cite{simon2017hand}. In order to filter out badly detected keypoints, we set all keypoints with a confidence score lower than $0.1$ as 0 for BRED and lower than $0.3$ for the GEMEP database. \textcolor{black}{These thresholds result in a percentage of approximately 70\% valid joints in each database.} The total size of the input vector for the body expression recognition branch is $134$: 25 2D keypoints of the skeleton and 21 2D keypoints for each hand.

\begin{table}[]
\centering
\begin{tabular}{|l|c|}
\hline
  \multicolumn{2}{|c|}{Video}  \\ \hline
               Method  &  ACC  \\
\hline 
  Body br. (TCN) & 0.31  \\
 Body br. (LSTM) & 0.28   \\ 
 Body br. (GTAP) & \textbf{0.34}    \\  \hline
Face br. & 0.43      \\
Whole Body br.  & \textbf{0.51}    \\ 
Human Baseline & 0.47 \cite{banziger2012introducing} \\
\hline

\hline
\end{tabular}
\begin{tabular}{|l|c|}
\hline
\multicolumn{2}{|c|}{Frame} \\ \hline
               Method  &  ACC \\
\hline 
 Body br.  & 0.23  \\
 Face br. & 0.21  \\ 
 Whole Body br.  & \textbf{0.33}   \\  
\hline

\end{tabular}

\caption{Accuracy results for the body, face, and whole body branch on the GEMEP database (12 classes).}
\label{tab:gemep_results}
\vspace{-0.5cm}
\end{table}

\setlength{\extrarowheight}{1pt}
\begin{table*}[t]
\centering
\begin{tabular}{p{10pt}|l|cc|cc|cc}
              & Label & \multicolumn{2}{c|}{$y$ (6 classes)}  & \multicolumn{2}{c|}{$y^f$ (7 classes)} & \multicolumn{2}{c}{$y^b$ (7 classes)} \\ \hline
             &  &  F1 & ACC & F1 & ACC & F1 & ACC \\
\hline

\multirow{3}{*}{\begin{turn}{90} SEP \end{turn}} &Body br. & 0.30 (0.29) & 0.35 (0.33) & - & - & 0.34 (0.48)  & 0.37 (0.46)     \\
& Face br. & 0.60 (0.62) & 0.65 (0.65)   & 0.54(0.61) & 0.59 (0.63) & - & -  \\
& Sum Fusion & 0.62 (0.64) & 0.65 (0.66) & -  &  -  & - & -\\ \hline

\multirow{3}{*}{\begin{turn}{90} \end{turn}}  & Joint-1L & 0.66 (0.67) & 0.67 (0.67) & - & - & - & -  \\ \hline

\multirow{3}{*}{\begin{turn}{90} \multirowcell{2}{HMT-3a}   \end{turn}}  & Body br. & 0.30 (0.30) & 0.34 (0.33) & - & - & 0.32 (0.44) & 0.36 (0.44)     \\
&Face br.  &  0.58 (0.61) & 0.65 (0.66) & 0.53 (0.59) & 0.60 (0.64)  & -  & - \\
&Fusion    & 0.67 (0.69) & 0.69 
(0.70) & -  &  - & - & -\\ \hline

\multirow{3}{*}{\begin{turn}{90} HMT-3b \end{turn}}  & Body br. & 0.29 (0.29) & 0.33 (0.32) & - & - & 0.35 (0.47) & 0.38(0.46) \\
& Face br. & 0.57 (0.60) & 0.64 (0.66) & 0.54 (0.59) & 0.60 (0.65) & - & - \\
& Whole body br.   & 0.65 (0.67) & 0.68
(0.69) & -  &  - & - & -\\ \hline

\multirow{3}{*}{\begin{turn}{90} HMT-4 \end{turn}}  & Body br. & 0.30 (0.30) & 0.34 (0.32) & - & - & 0.32 (0.44) & 0.36(0.44) \\
& Face br. & 0.57 (0.60) & 0.64 (0.66) & 0.53 (0.59) & 0.59 (0.64) & - & - \\
& Fusion   & \textbf{0.70} (\textbf{0.71}) & \textbf{0.72} 
(\textbf{0.72}) & -  &  - & - & -\\ \hline

\end{tabular}
\caption{Detailed results on the BRED database for various configurations of the HMT network. \textcolor{black}{Numbers outside parentheses report balanced scores and inside parentheses unbalanced scores. The highest achieved scores when evaluating against whole body labels are shown in bold.}}
\label{tab:babyrobot_results}
\vspace{-0.5cm}
\end{table*}

\subsection{Exploratory Results on the GEMEP Database}
The GEMEP database includes videos of 10 adult actors performing 17 different emotions: Admiration, Amusement, Anger, Anxiety, Contempt, Despair, Disgust, Fear, Interest, Irritation, Joy, Pleasure, Pride, Relief, Sadness, Surprise, and Tenderness. In this work we use the core set of the database that includes the first 12 of the aforementioned emotions. 

We use 10-fold leave-one-subject-out cross-validation and repeat the process for 10 iterations, averaging the scores in the end. For all different evaluation setups, we train for 200 epochs, reducing the learning rate by a factor of 10 at 150 epochs. We report Top-1 accuracy for several experimental setups in Table \ref{tab:gemep_results}. For the body expression recognition branch we compare three different implementations: a) the implementation with global temporal average pooling (GTAP) using a hidden FC layer of 256 neurons with ReLU activation, b) a temporal convolutional network (TCN) \cite{bai2018empirical} with 8 temporal convolutional residual blocks, 128 channels and kernel size 2, and c) a bidirectional long short-term memory network (LSTM) \cite{lstm} with 100 hidden units and two layers preceded by an FC layer of 128 neurons with activation. \textcolor{black}{For both TCN and LSTM we average the outputs over all time steps. In the first part of the table we observe that GTAP (shown in bold) achieves the highest accuracy (0.34) although it's a much simpler method. We believe that due to the small amount of data the methods focus only on certain representative postures that occur during the expression of emotions and ignore sequential information. As a result, the LSTM and TCN cannot outperform the DNN combined with GTAP, and would require a larger database in order to accurately capture temporal information. The face branch achieves a higher accuracy score (0.43) than the body branch (0.34), which is an expected result. Our main observation is the fact that the whole body emotion recognition branch (with the GTAP implementation) (shown in bold) achieves a significant improvement over the face branch baseline (an absolute 8\% improvement, from 43\% to 51\%).}

In Table \ref{tab:gemep_results} we also include experiments at the frame level, where we take only the middle frame of each video sequence and skip the temporal pooling structures in each branch. \textcolor{black}{We observe that again the whole body emotion recognition branch (in bold) yields a large performance boost over the facial branch (from 21\% to 33\%), as well as the significance of applying temporal pooling over all video frames.}

\begin{figure}[t]
  \includegraphics[width=\linewidth]{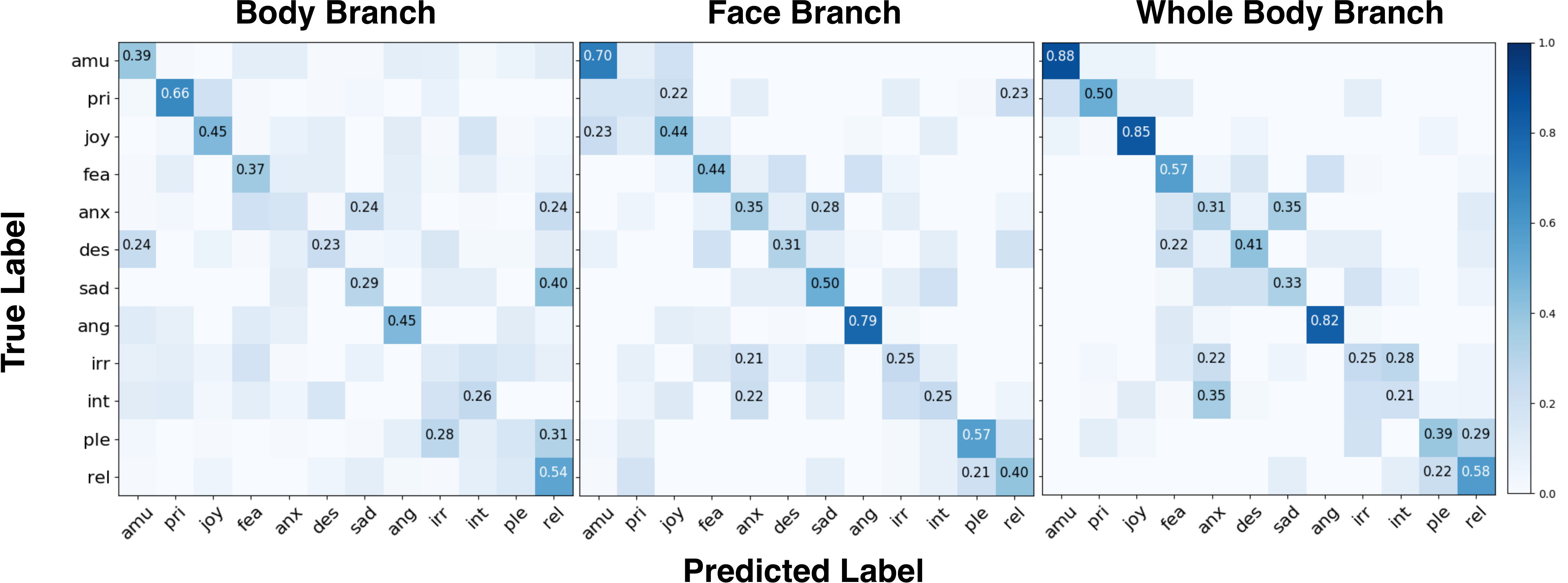}
    \caption{Confusion matrices for the face, body and whole body branches of HMT in the GEMEP corpus.}
    \label{fig:gemep_confusion_matrices}
      \vspace{-0.5cm}
\end{figure}

\textcolor{black}{Emotion specific details can be seen in the confusion matrices of Figure \ref{fig:gemep_confusion_matrices}. We show the confusion matrices for the separately trained body, face, and whole body branches.} We can see that in cases such as pride, the body branch is much more efficient in recognizing the emotion, as opposed to the face branch, a result which is also in line with \cite{tracy2004show}. In other emotions such as joy and anger, combination of face and body posture results in a higher accuracy. There are also emotions for which the body branch fails to learn any patterns such as anxiety or pleasure. In these cases, the whole body branch achieves a lower accuracy than the face branch.

\begin{figure*}[t]
\centering
   \includegraphics[width=1\linewidth]{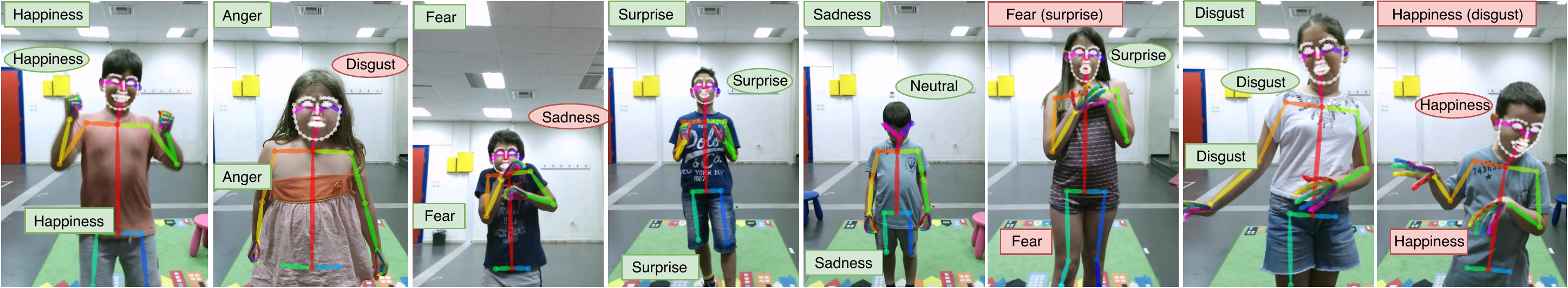}
\caption{Example results of whole body affect recognition. Captions on top of each image denote the final predictions while oval shapes denote predictions of the face branch and rectangle shapes denote predictions of the body branch. Green color shapes denote a correct prediction, whereas red color shapes denote an incorrect prediction. If the final predicted label is wrong then inside parenthesis we include the correct label.}
    \label{fig:results}
\end{figure*}

\begin{figure}[t]
   \includegraphics[width=\linewidth]{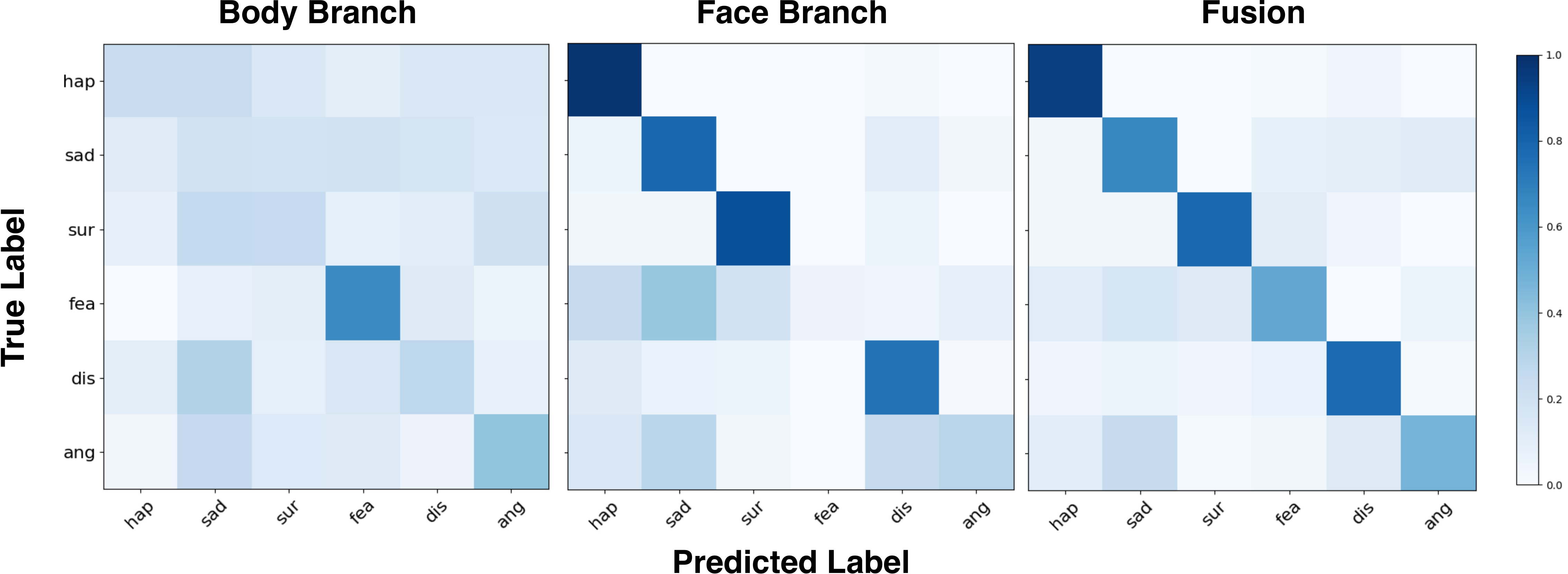}
    \caption{Confusion matrices of the body, face, and fusion branches of HMT-4, against whole body labels $y$ on the BRED database.}
    \label{fig:babyrobot_confusion_matrices}
    \vspace{-0.5cm}
\end{figure}

\subsection{Results on the BabyRobot Emotion Database}
For BRED we follow the exact same procedure as with the GEMEP database: training for 200 epochs, reducing the learning rate by a factor of 10 at 150 epochs, and 10-fold cross validation for 10 iterations. For the 10-fold cross validation, we ensure that each subject (30 in total) does not appear in both the training and test set of the same split.
Because the database is highly unbalanced, especially for the body labels, we report results in balanced and unbalanced F1-score and accuracy. Due to this imbalance we also use a balanced cross entropy loss for $\mathcal{L}^{b}$, since the amount of instances labeled as neutral are much larger than the emotion instances. We also note that for BRED, the annotations $y^{f}$ and $y^{b}$ include 7 classes (all emotions plus neutral), while the whole body annotation $y$ includes 6 classes (all emotions).

We report our results in Table \ref{tab:babyrobot_results}. The column labeled with $y$ reports the metrics on the whole body labels, while columns $y^{f}$ and $y^{b}$ report results on the hierarchical face and body labels, respectively. For calculating the metrics of the face and body branches against $y$, we ignore the scores of the ``neutral" label.
Numbers outside parentheses report balanced scores and inside parentheses unbalanced scores. \textcolor{black}{The highest achieved scores when evaluating against whole body labels are shown in bold.}

Table \ref{tab:babyrobot_results} contains results of $5$ different methods:  \textit{SEP} denotes independent training of the body and face branch using their corresponding labels. 
\textit{Joint-1L} denotes training of the whole body emotion branch and only using the $\mathcal{L}^{w}$ loss.
\textit{HMT-3a} denotes joint training of the hierarchical multi-label training network, if we omit the branch of the whole body emotion recognition, i.e., with the losses $\mathcal{L}^d$, $\mathcal{L}^f$, and $\mathcal{L}^b$. \textit{HMT-3b} denotes joint training of the three losses: $\mathcal{L}^b$, $\mathcal{L}^f$, and $\mathcal{L}^w$, by omitting the final score fusion. Finally, \textit{HMT-4} denotes the joint training with all four losses of the HMT network. In the methods that include the fusion branch, we obtain the final prediction by the scores of the fusion $s^d$. In the case of HMT-3b, where we omit the final fusion, we obtain the final whole body label prediction by the whole body branch.

\textcolor{black}{Our initial observation is the fact that the combination of body posture and facial expression results in a significant improvement over the facial expression baselines, for all different methods. Secondly, we see that HMT-4 achieves the highest scores for all metrics (0.70 balanced F1-score and 0.72 balanced accuracy), across all methods, as far as the whole body emotion label is concerned, while HMT-3a and HMT-3b exhibit similar performance (0.67 and 0.65 balanced F1-score, respectively) that is also comparable to the separate training of the body and face branches and their combination with post-process sum-based fusion (0.62).}

We remind that $y^{f}$ and $y^{b}$ have one more class than $y$ (neutral), which is why the scores appear lower for the face branch in the $y^{f}$ column. This is not the case for the body branch, due to the fact that $y^{b}$ and $y$ are different by a lot more labels (99), while $y^{f}$ and $y$ differ in only $37$ labels.

In Figure \ref{fig:results} we present several results (both correct and incorrect recognitions) of our method, while in Figure \ref{fig:babyrobot_confusion_matrices} we also depict the confusion matrices for the body, face, and fusion predictions when fared against the whole body labels $y$. We observe that generally, due to the fact that children in BRED relied more on facial expressions than bodily expressions (as it was observed in Table \ref{tab:stats-ann}), only including the body branch in a system would result in low performance. We also observe that the face branch achieves low recognition rates for fear and anger. However, fusing the two using the HMT network results in a model that can reliably recognize all emotions.


\section{Conclusions}
\label{sec:6}
In this work we proposed a method for automatic recognition of affect that combines whole body posture and facial expression cues in the context of CRI. CRI presents a challenging application that requires leveraging body posture for emotion recognition and cannot rely only on facial expressions. The proposed method can be trained both end-to-end, as well as individually, and leverages multiple hierarchical labels providing computational models that can be used jointly and individually.

We performed an extensive evaluation of the proposed method on the BabyRobot Emotion Database that features whole body emotional expressions of children during a CRI scenario. 
Our results show that fusion of body and facial expression cues can be used to significantly improve the emotion recognition baselines that are based only on facial expressions, and that 2D posture can be used with promising results for emotion recognition. We also show that hierarchical multi-label training can be exploited for improving system performance.

We believe our research shows promising results towards establishing body posture as a necessary direction for emotion recognition in human-robot interaction scenarios, and highlights the need for creating large-scale whole body emotional expression databases.









\bibliographystyle{IEEEtran}
\bibliography{egbib}


\addtolength{\textheight}{-12cm}   

\end{document}